\documentclass[letterpaper]{article} 
\usepackage{aaai2026}  
\makeatletter
\def\@copyrightspace{\relax}
\makeatother
\usepackage{times}  
\usepackage{helvet}  
\usepackage{courier}  
\usepackage[hyphens]{url}  
\usepackage{graphicx} 
\usepackage{natbib}  
\usepackage{caption} 
\usepackage{algorithm}
\usepackage{algorithmic}
\usepackage{booktabs}
\usepackage{amsmath} 
\usepackage{times}
\usepackage{helvet}
\usepackage{courier}
\usepackage{xcolor}
\usepackage{bm}

\usepackage{newfloat}
\usepackage{listings}
\DeclareCaptionStyle{ruled}{labelfont=normalfont,labelsep=colon,strut=off} 
\floatstyle{ruled}
\newfloat{listing}{tb}{lst}{}
\floatname{listing}{Listing}
\title{SortBlock: Similarity-Aware Feature Reuse for Diffusion Model}
\author{
    Hanqi Chen\textsuperscript{\rm 1}\thanks{Equal Contribution},
    Xu Zhang\textsuperscript{\rm 2}\footnotemark[1],
    Xiaoliu Guan\textsuperscript{\rm 3},
    Lielin Jiang\textsuperscript{\rm 2},
    Guanzhong Wang\textsuperscript{\rm 2}\thanks{Corresponding Author},
    Zeyu Chen\textsuperscript{\rm 2},  
    Yi Liu\textsuperscript{\rm 2}
}

\affiliations{

    \textsuperscript{\rm 1}International Joint Innovation Center, The Electromagnetics Academy at Zhejiang University\\
    \textsuperscript{\rm 2}PaddlePaddle Team, Baidu Inc. \\
    \textsuperscript{\rm 3}School of Computer Science, Wuhan University\\

}

\usepackage{bibentry}

\begin{document}

\maketitle

\begin{abstract}
Diffusion Transformers (DiTs) have demonstrated remarkable generative capabilities, particularly benefiting from Transformer architectures that enhance visual and artistic fidelity. However, their inherently sequential denoising process results in high inference latency, limiting their deployment in real-time scenarios. Existing training-free acceleration approaches typically reuse intermediate features at fixed timesteps or layers, overlooking the evolving semantic focus across denoising stages and Transformer blocks.To address this, we propose SortBlock, a training-free inference acceleration framework that dynamically caches block-wise features based on their similarity across adjacent timesteps. By ranking the evolution of residuals, SortBlock adaptively determines a recomputation ratio, selectively skipping redundant computations while preserving generation quality. Furthermore, we incorporate a lightweight linear prediction mechanism to reduce accumulated errors in skipped blocks. Extensive experiments across various tasks and DiT architectures demonstrate that SortBlock achieves over 2 times inference speedup with minimal degradation in output quality, offering an effective and generalizable solution for accelerating diffusion-based generative models.
\end{abstract}
\begin{links}\link{Project Page}{https://sortblock.github.io}\end{links}

\section{Introduction}

Diffusion models have been widely recognized as a key breakthrough in image and video generation, owing to their powerful generative capabilities. Recently, there has been growing interest in shifting the architecture of diffusion models from U-Net to Transformer-based designs such as Diffusion Transformers (DiTs)\cite{peeblesScalableDiffusionModels2023}. These refined architectures not only produce visually and artistically compelling outputs but also demonstrate superior scaling behavior.

Despite their impressive performance, DiTs remain constrained by slow inference speeds due to their inherently sequential denoising process, limiting their applicability in real-time scenarios. The core bottleneck stems from the iterative nature of the reverse process, which hinders parallel decoding. Moreover, as model sizes grow and higher-resolution or longer-duration video generation becomes more desirable, the inference latency becomes even more problematic.

To accelerate the visual generation process, prior work has explored distillation-based approaches\cite{geSenseFlowScalingDistribution,yinOnestepDiffusionDistribution2024,yinImprovedDistributionMatching2024} and post-training techniques. However, these methods typically require additional training, incurring substantial computational costs and data demands. An alternative line of research focuses on caching mechanisms, which aim to accelerate inference without retraining. Such methods leverage the observation that model outputs across adjacent timesteps during denoising tend to be similar, enabling unified caching strategies to reduce redundant computation. Some approaches further propose predictive reuse to improve accuracy over direct caching, while others investigate output patterns across blocks to devise reuse policies.

In this work, we aim to accelerate DiTs by minimizing redundant computation, a plug-and-play paradigm that can generalize across architectures and tasks. While feature redundancy has been widely acknowledged in visual tasks, and recent works~\cite{xieSANA15Efficient2025,xieSANAEfficientHighResolution2024} have observed its presence in diffusion denoising, redundancy within DiTs and how to mitigate it remains underexplored.

To address this gap, we revisit the feature distances between DiT blocks(as shown in Fig.~\ref{fig:3}), and propose a training-free acceleration method called SortBlock, which predicts block-level changes in the next timestep and exploits high-similarity transitions to avoid redundant computation. Prior reuse methods~\cite{yuABCacheTrainingFreeAcceleration2025,shenLazyDiTLazyLearning2025} often adopt uniform strategies without accounting for multi-scale or varying semantic similarities. Excessive reuse can introduce compounding errors, leading to distorted structures and misalignment with prompts.

In the denoising process, structural content is typically formed at early stages with higher noise levels, while textures and details emerge at later stages with lower noise levels. Moreover, caching deeper (later) DiT blocks tends to favor contour generation, whereas caching shallower (earlier) blocks is more beneficial for preserving fine-grained details. Therefore, applying a static reuse policy derived from local patterns to the entire process may be too arbitrary. We argue that designing stage-specific reuse strategies is both more principled and more effective in capturing redundancy.

SortBlock divides the denoising process into multiple strategy intervals. For each interval, it ranks DiT blocks based on the similarity of residual changes between the first two timesteps and selectively recomputes blocks with the lowest similarity. To address the issue of growing errors from direct feature reuse, we further introduce a simple effective linear extrapolation mechanism that predicts the next-step feature and reduces reconstruction error with negligible overhead. Despite its simplicity, this strategy enables more aggressive caching while maintaining generation quality.

We evaluate SortBlock on the Flux.1-dev\cite{flux2024}, Wan2.1\cite{wanteamWanOpenAdvanced2025a} and HunyuanVideo\cite{kongHunyuanVideoSystematicFramework2025} baseline. Experimental results show that our training-free method achieves over 2× inference speedup without sacrificing perceptual quality. Moreover, SortBlock consistently outperforms other training-free acceleration baselines.


\section{Related Work}
\textbf{Diffusion Transformer.} Diffusion models have emerged as key players in the generative modeling landscape due to their impressive capabilities. Traditionally, U-Net-based diffusion models~\cite{zhangDiffusionModelNoiseAware2025,rombachHighResolutionImageSynthesis2022} have demonstrated strong performance across various applications, including image and video generation~\cite{wanteamWanOpenAdvanced2025a,maLatteLatentDiffusion2025}. Recently, a growing body of work has transitioned to Transformer-based architectures. This architectural shift has led to significant improvements in generating visually and artistically compelling content, while also better adhering to scaling laws. Moreover, DiTs show promising potential in effectively integrating and generating multimodal content. However, their real-time applicability remains limited by the inherently iterative nature of the diffusion process.

\textbf{Acceleration of Diffusion Models.} To accelerate the inference process of diffusion models, extensive efforts have been made, which can generally be categorized into two paradigms: reducing the number of sampling steps and reducing the per-step computational cost. The former typically involves designing faster samplers~\cite{songDenoisingDiffusionImplicit2022a,luDPMSolverFastODE2022a,zhangAdaDiffAdaptiveStep2024,zhaoUniPCUnifiedPredictorCorrector2023}, or performing step-wise distillation~\cite{yinOnestepDiffusionDistribution2024,mengDistillationGuidedDiffusion2023}. The latter focuses on model-level optimization techniques such as distillation~\cite{geSenseFlowScalingDistribution,yinImprovedDistributionMatching2024,starodubcevScalewiseDistillationDiffusion2025}, pruning~\cite{xieSANA15Efficient2025,fangStructuralPruningDiffusion2023}, or reducing redundant computations~\cite{liuDLLMCacheAcceleratingDiffusion,maDeepCacheAcceleratingDiffusion2023,zhangBlockDanceReuseStructurally2025,liuTimestepEmbeddingTells2025,yuABCacheTrainingFreeAcceleration2025,lvFasterCacheTrainingFreeVideo2025}. Some prior studies have identified feature redundancy in U-Net-based diffusion models~\cite{chen$D$DiTTrainingFreeAcceleration2024}, however, their coarse-grained feature reuse strategies often include low-similarity features, leading to structural distortions and misalignment between text and generated images. In contrast, we investigate feature redundancy specifically in DiTs and propose reusing structurally similar spatiotemporal features to achieve acceleration while maintaining high consistency with the baseline model outputs.
\section{Methodology}
\subsection{Preliminaries}
\textbf{Forward and Reverse Process in Diffusion Models.} Diffusion models gradually inject noise into data and learn to reverse this process in order to generate clean data from noise. In this work, we focus on the latent-space formulation of noise injection and denoising introduced by ~\cite{rombachHighResolutionImageSynthesis2022}, where the diffusion process is performed in the latent space. In the forward process, the posterior distribution of the noisy latent variable $z_t$ at timestep t admits a closed-form expression:
\begin{equation}
q(z_t|z_0) = \mathcal{N}(z_t; \sqrt{\overline{\alpha}_t}z_0, (1 - \overline{\alpha}_t)I),
\end{equation}
the noise variance sequence is defined as $\bm{\alpha_t = \prod_{i=0}^t (1 - \beta_i)}$, where $\bm{\beta_i \in (0,1)}$. The reverse process, which entails generating data from noise, constitutes a pivotal component within the diffusion model framework. Upon training the diffusion model $\bm{\epsilon_\theta(z_t, t)}$, the conventional sampler DDPM progressively denoises $\bm{z_t \sim \mathcal{N}(0, I)}$ across $\bm{T}$ steps. Alternatively, faster samplers akin to DDIM can expedite this procedure utilizing the subsequent formula:

\begin{equation}
\begin{aligned}
\mathbf{z}_{t-1} = \sqrt{\alpha_{t-1}} \left( 
    \frac{\mathbf{z}_t - \sqrt{1 - \alpha_t} \, \epsilon_\theta(\mathbf{z}_t, t)}{\sqrt{\alpha_t}} 
\right) \\
\quad + \sqrt{1 - \alpha_{t-1} - \sigma_t^2} \cdot \epsilon_\theta(\mathbf{z}_t, t) 
+ \sigma_t \epsilon_t.
\end{aligned}
\label{eq:2}
\end{equation}

During this denoising phase, the model initially generates the coarse structure of the image and subsequently refines it by incorporating textures and detailed information.

\textbf{Features In The Transformer.} The number of denoising steps directly correlates with the number of network forward passes in the DiT architecture, which is typically composed of multiple stacked blocks. Each block sequentially computes its output based on the input from the previous one. The shallow blocks, being closer to the input, tend to capture global structures and coarse layouts of the data. In contrast, deeper blocks, located closer to the output, progressively refine fine details, contributing to outputs that are both realistic and visually compelling ~\cite{parkHowVisionTransformers2022,wuVisualTransformersWhere2021}.

\begin{figure}[t]
\centering
\includegraphics[width=0.45\textwidth]{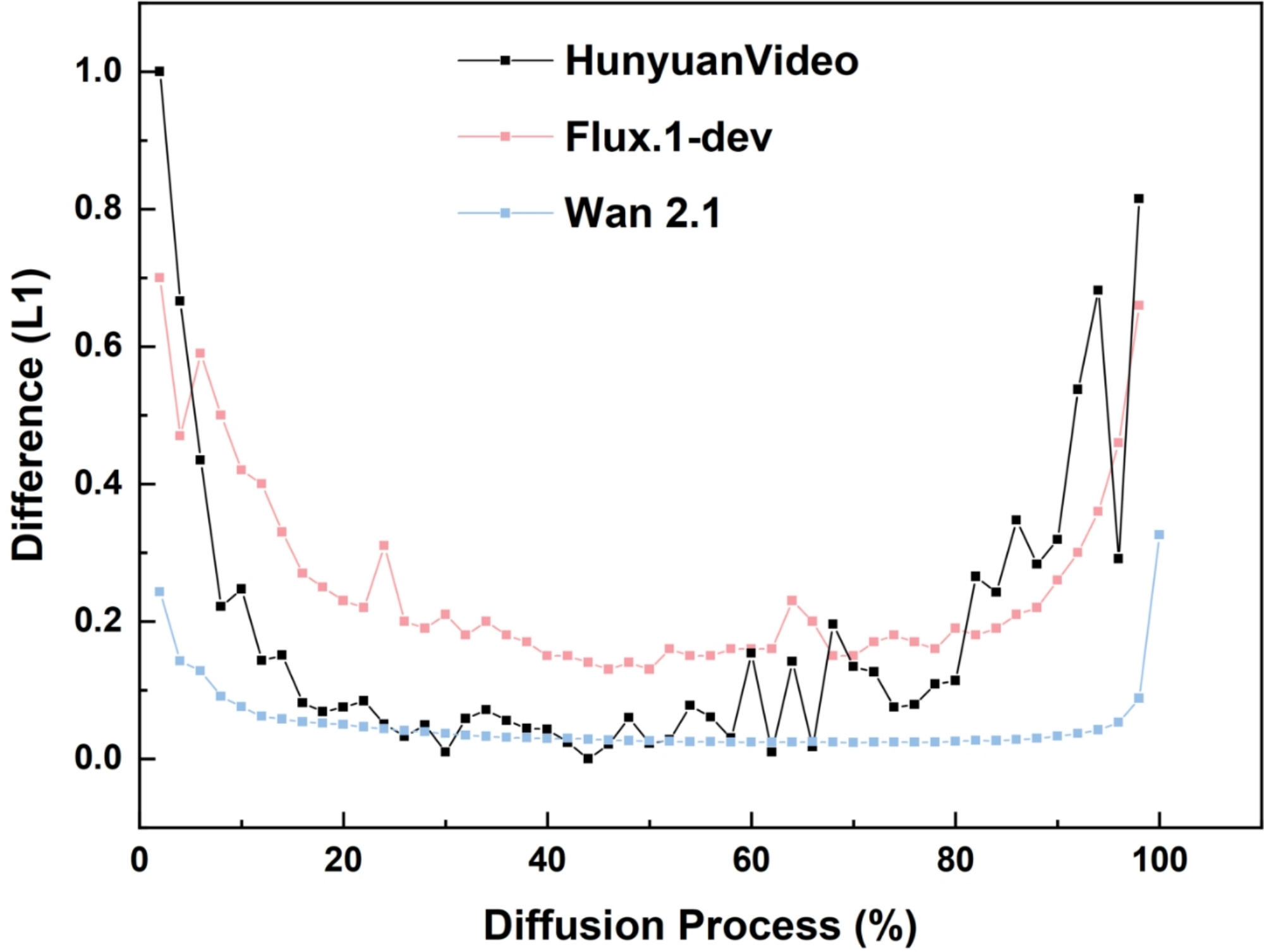} 
    \captionof{figure}{Visualization of output differences in consecutive timesteps of Wan2.1, HunyuanVideo and Flux.1-dev.} 
    \label{fig:L1}
\end{figure}

\begin{figure*}[t]
\centering
\includegraphics[height=8.5cm,width=1\textwidth]{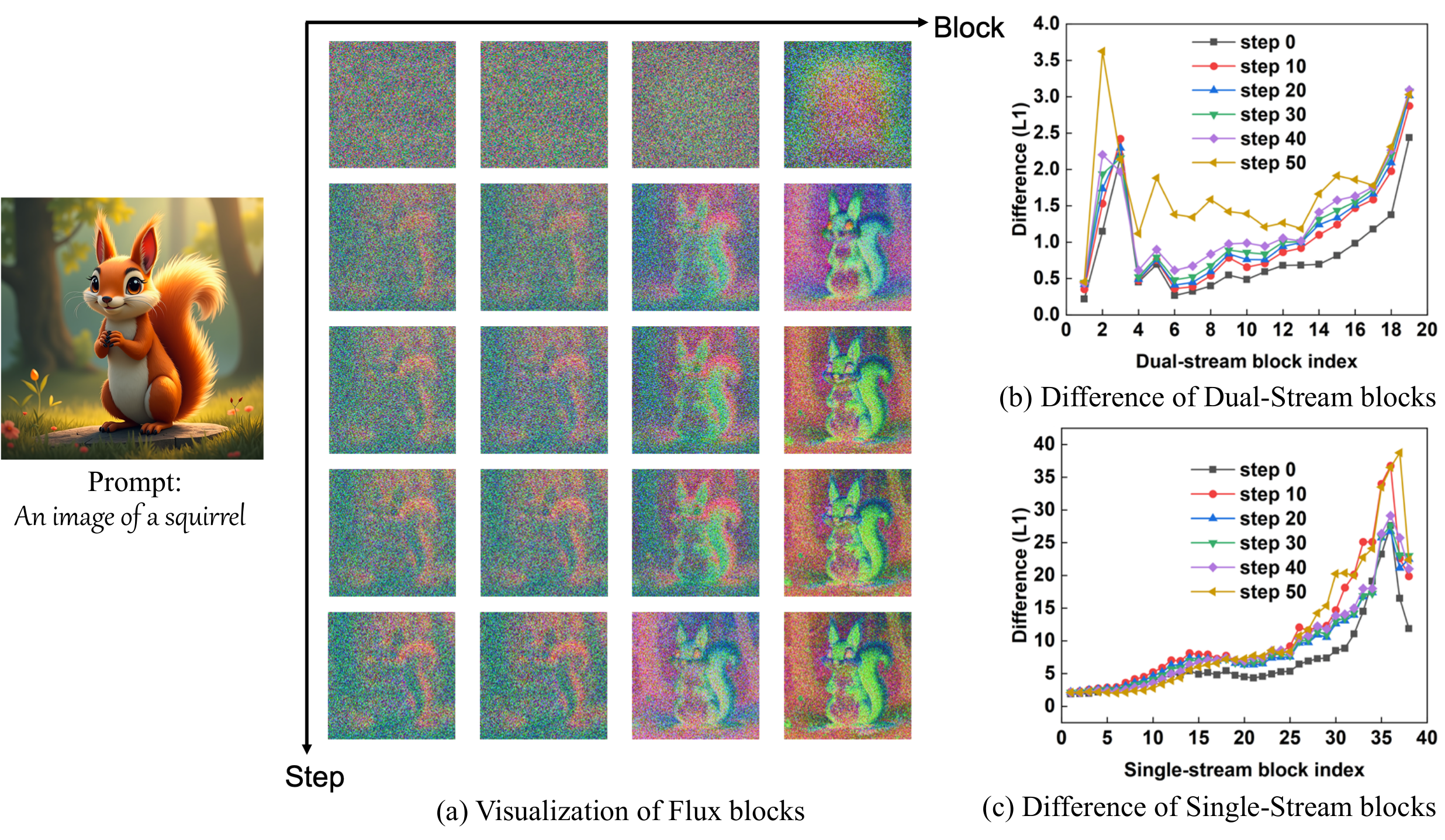} 
    \captionof{figure}{\textbf{Visualization of Flux.1-dev blocks.} (a) Visualization of block dynamics as the network depth and timesteps increase. (b) L1 distance variations between dual-stream attention blocks in Flux.1-dev. (c) L1 distance variations between single-stream attention blocks in Flux.1-dev.} 
    \label{fig:3}
\end{figure*} 
\subsection{Analysis}
To investigate the correlation between DiT blocks and adjacent timesteps, we conduct an in-depth analysis of their behavior throughout the diffusion process.

\textbf{Difference with Timesteps.}  To better explore variations across timesteps at different denoising stages, we quantify the output differences across all timesteps during the diffusion process, as illustrated in Fig.~\ref{fig:L1}. Our analysis reveals that approximately 70\% of the diffusion steps—primarily in the middle region—exhibit low variation in attention outputs. This indicates a high degree of redundancy and serves as the main target for our caching strategy. In contrast, the early and late stages show significantly larger variations, suggesting that these regions require higher recomputation ratios—or even full recomputation—in order to minimize the discrepancy between accelerated and original model outputs.

\textbf{Difference with Blocks.}  In an ideal scenario, if we could obtain in advance the variations of each DiT block at the next time step, we could directly measure the differences between the variations of adjacent time step DiT blocks and decide whether to recompute these DiT blocks based on these differences. In this paper, we use cosine similarity as the metric. For instance, the cosine similarity for the variations of the same DiT block across adjacent time steps is calculated as follows:

\begin{equation}
\mathbf{S}_N = \frac{(\Delta_k^N)^T \Delta_{k+1}^N}{\|\Delta_k^N\| \|\Delta_{k+1}^N\|},
\label{eq:3}
\end{equation}
Where $\Delta_k^N$ represents the input-output variation of the $N$-th DiT block at the $K$-th step, and $\Delta_{k+1}^N$ represents the input-output variation of the $N$-th DiT block at the $(K+1)$-th step. A value of $\mathbf{S}_N$ closer to 0 indicates greater differences between the two DiT blocks, suggesting they should be recomputed. Conversely, an $\mathbf{S}_N$ closer to 1 implies higher similarity, allowing for the reuse of information from the previous time step to accelerate model inference speed. Therefore, Eq.~\ref{eq:3} serves as the criterion to define and judge whether a DiT block needs to be recomputed.

\textbf{Feature Analysis.} However, in most cases, the model's output cannot be obtained in advance, rendering the aforementioned methods infeasible. To address this issue, an intuitive approach is to design a caching strategy by effectively estimating the input-output changes of the DiT blocks. Taking the Flux.1-dev model as an example, as shown in Fig.~\ref{fig:3} (b)(c), the vertical axis represents the L1 distance between the input and output of each block. In the middle region of the dual-stream block and most preceding single-stream blocks, the feature representation exhibits a smooth evolution as the layer depth increases. Based on this insight, we aim to make a simple prediction of the input-output changes for the DiT blocks in the next time step of the model to satisfy the similarity criterion outlined in Eq.~\ref{eq:3}.

\begin{figure*}[t]
\centering
    \includegraphics[height = 9cm,width=1\textwidth]{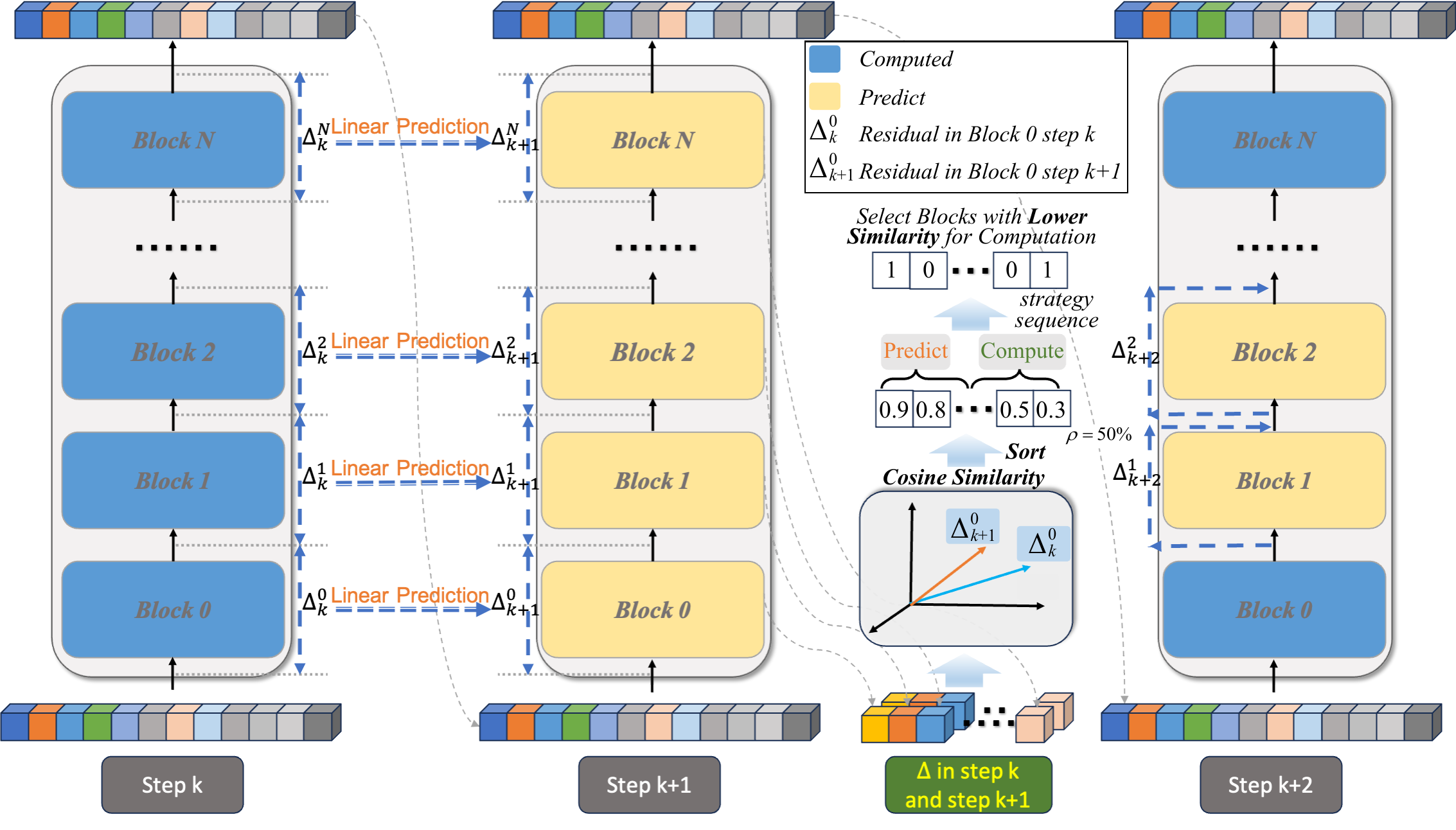} 
    \captionof{figure}{\textbf{The SortBlock pipeline.} Block features are adaptively updated based on their similarity between fully computed and predicted counterparts. Specifically, blocks with low similarity are recomputed, while the others are efficiently approximated via prediction.} 
    \label{fig:4}
\end{figure*} 
\subsection{Training-Free Acceleration Approach} To alleviate the low inference efficiency problem of DiTs, we introduce SortBlock, a training-free caching framework. Diffusion models operate by gradually denoising images step by step based on time steps, however, this evolutionary process exhibits sparsity because the changes in features between consecutive steps during intermediate denoising are often highly similar. As illustrated in Fig.~\ref{fig:3}, this sparsity suggests that recalculating all features at every step is generally unnecessary.

To leverage this sparsity, SortBlock selectively updates only a small subset of DiT blocks that exhibit the most significant changes between adjacent steps. The challenge lies in efficiently and accurately identifying these DiT blocks. As shown in Fig.~\ref{fig:4}, we compute the cosine similarity between the input-output changes of DiT blocks at step K and step K+1. By utilizing the cosine similarity between the input-output changes of each DiT block, we identify those DiT blocks with the most substantial changes. Only these selected DiT blocks undergo complete feature recomputation and cache updates.

Calculating the cosine similarity requires features from two adjacent time steps, and full computation of these features would incur high time costs, thus limiting the upper bound of the strategy's speed. By leveraging the continuous nature of feature trajectories, we employ a predictive caching method that estimates intermediate features across time steps.

\textbf{Linear Prediction Method.} Unlike the approach of directly copying features, we cache both the feature values and their time differences: $ C(x_t^N) := \{ \mathcal{F}(x_t^N), \Delta \mathcal{F}(x_t^N) \} $. This enables us to predict the feature trajectory at time $ t - k $ using the formula:

\begin{equation}
\mathcal{F}_{\text{pred}}(x_{t-k}^N) = \mathcal{F}(x_t^N) + \frac{\mathcal{F}(x_t^N) - \mathcal{F}(x_{t+L}^N)}{L} k,
\label{eq:4}
\end{equation}
where $ \mathcal{F}(x_t^N) $ and $ \mathcal{F}(x_{t+L}^N) $ represent the eigenvalues calculated at the complete time steps $t$ and $t+L$, respectively. $L$ denotes the time step interval. This first-order approximation effectively captures the linear trend of the feature trajectory, thereby significantly enhancing the accuracy of direct feature reuse. Notably, our feature prediction mechanism is not restricted to calculating subsequent time steps or for similarity computations, instead, it replaces all components that require cached and reused feature values with linear predictions.

The caching strategy is governed by two hyperparameters: the \textbf{policy refresh interval} $ K $ and the \textbf{self-adaptive update ratio} $ \rho \in [0,1] $. The inference process typically involves these components:

\textbf{Initialization:} When $k \equiv 0 \pmod{K}$, all features are computed, and the input-output changes $\Delta_k^N$ for all DiT blocks are stored.

\textbf{Linear Prediction:} When $k \equiv 1 \pmod{K}$, based on the feature values obtained from the previous time step, all features for the current time step are predicted using Eq.~\ref{eq:4}, and the input-output changes $\Delta_{k+1}^N$ for all DiT blocks are stored.

\textbf{Identify Changes:} The cosine similarity $s_N$ between the current input-output changes $\Delta_{k+1}^N$ and the previous changes $\Delta_k^N$ is calculated for each DiT block.

\textbf{Select DiT Blocks:} The $[\rho N_{\text{max}}]$ DiT blocks with the lowest $s_N$ values are selected and marked as 1 in the strategy sequence $I$, with the remaining blocks marked as 0.

\textbf{Recompute and Predict:} Only the DiT blocks with $I_{\text{index}} = 1$ are recomputed; the rest are predicted using linear prediction.

\textbf{Update Cache:} The cached values are replaced with the newly computed values for the DiT blocks that were recomputed.

\textbf{Summary:} This adaptive update mechanism leverages temporal stability to minimize computational cost while maintaining generation quality through selective feature refreshment.

\begin{table*}[ht]
\centering

\resizebox{2\columnwidth}{!}{
\renewcommand{\arraystretch}{1}
\begin{tabular}{lccccccccc}
\hline
\textbf{Method} & \multicolumn{2}{c}{\textbf{Speed}} & \multicolumn{6}{c}{\textbf{Image Quality}} \\  \cmidrule(lr){2-3} \cmidrule(lr){4-9}
 & Speedup ↑  & Latency (s) ↓ & FID ↓ & IR ↑ & CLIP ↑ & SSIM ↑ & PSNR ↑ & LPIPs ↓ \\ 
\hline
Flux.1-dev, 50 steps & 1.0× & 23.23 & 70.59 & 1.064 & 0.332 & - & - & - \\
\hline
T-GATE (m=25) & 1.2× & 19.44 & \textbf{69.40} & 1.056 & \textbf{0.332} & 0.734 & 19.96 & 0.271 \\
PAB & 1.37× & 16.97 & 69.72 & 1.001 & 0.331 & 0.849 & 24.65 & 0.207 \\
TeaCache (rel=0.25) & 1.73× & 13.45 & 70.14 & 1.035 & 0.331 & 0.805 & 22.11 & 0.198 \\
Taylorseer (N=3 O=3) & 2.04× & 11.37 & 70.27 & 1.048 & 0.330 & 0.763 & 20.56 & 0.241 \\
\hline
SortBlock (K=5 (900,50)) & 2.00× & 11.60 & 70.47 & 1.067 & 0.327 & \textbf{0.952} & \textbf{31.49} & \textbf{0.0748} \\
SortBlock (K=9 (900,50)) & \textbf{2.39}× & \textbf{9.7} & 70.30 & \textbf{1.077} & 0.328 & 0.910 & 27.27 & 0.1256 \\
\hline
\end{tabular}
}
\caption{Text-to-image generation on Flux.1-dev.}
\label{tab:flux}
\end{table*}
\begin{figure*}
    \centering
\includegraphics[width=1\linewidth]{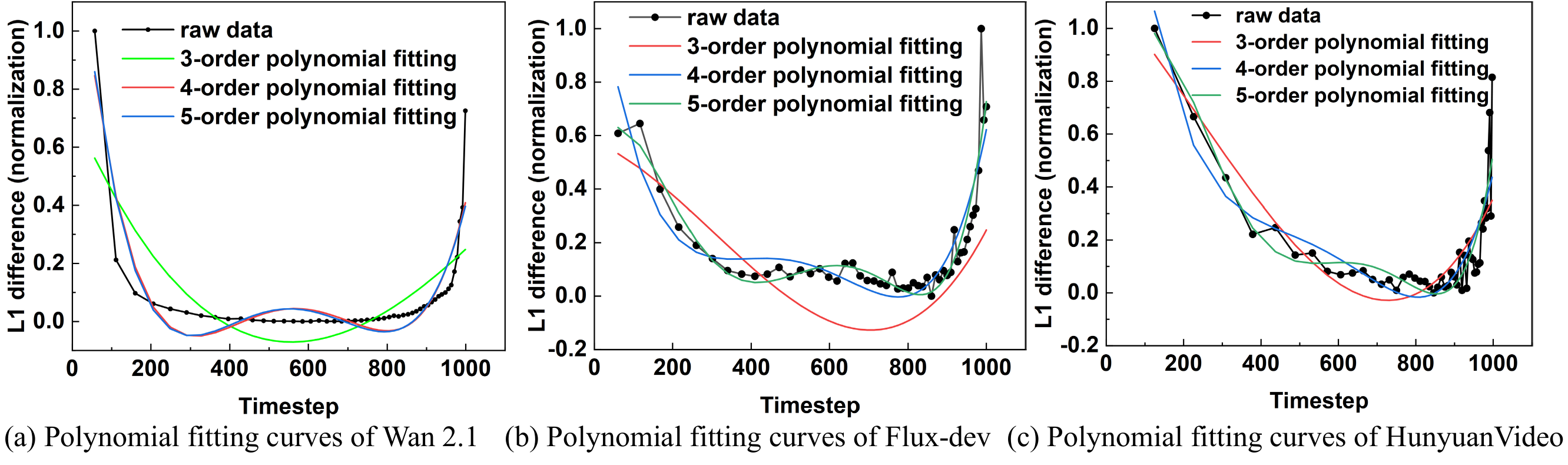}

   \caption{Correlation visualization between input and output differences across consecutive timesteps in Wan2.1, Flux.1-dev and HunyuanVideo. While raw data points deviate significantly from linearity, polynomial fitting improves the approximation accuracy.
}
   \label{fig:onecol}
\end{figure*} 

\section{Experiment}
\subsection{Settings}
\textbf{Base Models and Compared Methods.} To demonstrate the effectiveness of our approach, we apply our acceleration technique to both image and video generation models, including Flux.1-dev, Wan2.1~\cite{wanteamWanOpenAdvanced2025a} and HunyuanVideo~\cite{kongHunyuanVideoSystematicFramework2025}. To highlight its advantages, we compare our base models with several recent efficient video synthesis methods, such as PAB~\cite{zhaoRealTimeVideoGeneration2025}, T-GATE~\cite{liuFasterDiffusionTemporal2025}, TaylorSeer~\cite{liuReusingForecastingAccelerating2025} and TeaCache~\cite{liuTimestepEmbeddingTells2025}. Notably, T-GATE was originally designed for accelerating image synthesis, and PAB adapts it for video generation to enable fair comparison.

\textbf{Implementation Detail.} All image generation experiments are conducted on NVIDIA A800 80GB GPUs using the PaddlePaddle framework, while all video generation experiments are performed on NVIDIA H100 80GB GPUs with the same framework. FlashAttention~\cite{daoFlashAttentionFastMemoryEfficient2022} is enabled by default in all experiments.

For SortBlock, we apply caching and reuse primarily within the denoising range of 20\% to 100\% for both Flux.1-dev, Wan2.1 and HunyuanViedo. The recomputation ratio $\rho$ is defined as a timestep-dependent function, fitted to the curve of L1 distances between adjacent timesteps, as illustrated in Fig.~\ref{fig:onecol}. We compare polynomial fits of degree 3 to 5, and empirically choose a 5th-degree curve as the final formulation.

To make $\rho$ adjustable, we introduce a global scaling hyperparameter $\beta \in [0, 1]$, which controls the aggressiveness of the caching policy and allows tuning of the trade-off between speedup and generation quality.

\medskip
\begin{figure}[t]
  \centering
  \includegraphics[height=8cm, width=1\linewidth]{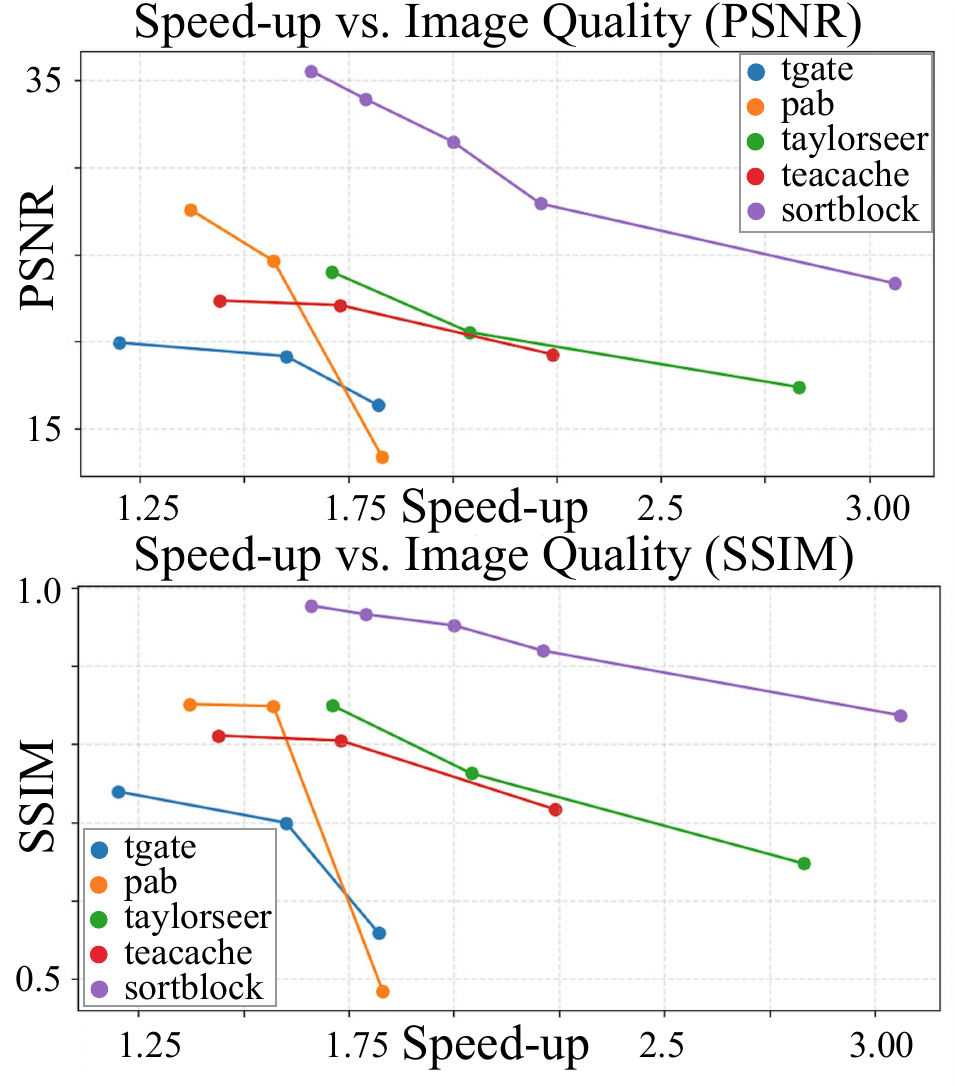} 
    \captionof{figure}{\textbf{Quality-latency comparison of cache method with different speed.} PSNR and SSIM comparisons of SortBlock, T-GATE, PAB, TaylorSeer, and TeaCache using Flux.1-dev. SortBlock achieves the best overall balance between visual quality and efficiency.} 
    \label{fig:5}
\end{figure}




\subsection{Main Results}
\textbf{Accelerate Flux.1-dev for Text-to-Image Generation.} We follow the evaluation protocol established in Flux.1-dev to assess our method on high-resolution text-to-image generation tasks. We report standard metrics including FID~\cite{heuselGANsTrainedTwo2018}, CLIP score~\cite{radfordLearningTransferableVisual2021}, Image Reward~\cite{xuImageRewardLearningEvaluating2023}, PSNR and SSIM~\cite{horeImageQualityMetrics2010} to comprehensively evaluate the visual quality and semantic alignment of generated images.

Inference efficiency is measured by the generation latency, defined as the average time to generate a single image or video on an NVIDIA A800 GPU. We validate the effectiveness of SortBlock on the COCO dataset. As shown in Table~\ref{tab:flux}, we compare our method against several recent acceleration baselines extended to the Flux.1-dev model, including PAB~\cite{zhaoRealTimeVideoGeneration2025}, T-GATE~\cite{liuFasterDiffusionTemporal2025}, TaylorSeer~\cite{liuReusingForecastingAccelerating2025} and TeaCache~\cite{liuTimestepEmbeddingTells2025}. 

For a fair comparison, we adapt each baseline to the Flux.1-dev backbone. Specifically, for PAB, we reduce the computational overhead by reconfiguring the broadcast mechanism. For T-GATE, the gating value is fixed to 25 throughout the denoising process. In TaylorSeer, we set the reuse factor to 3 and use a polynomial degree of 3. 

When the policy refresh interval $K=5$ and the policy is applied within timesteps $\in [50, 900]$, SortBlock achieves a $2\times$ speedup on Flux.1-dev while maintaining comparable image quality in terms of both visual aesthetics and prompt alignment. Notably, different speed–quality trade-offs can be flexibly controlled by tuning $K$ and the application interval of the caching strategy.

Compared to T-GATE and PAB, SortBlock achieves superior performance across all image quality metrics while maintaining comparable inference speed. Our method explicitly reduces redundant structural computations during the denoising stages by avoiding recomputation of highly similar block features, thereby minimizing quality degradation.

In contrast, T-GATE, TaylorSeer and PAB apply one or two fixed reuse strategies throughout the entire denoising process without considering the characteristics of different denoising stages. This often leads to suboptimal reuse policies across timesteps, resulting in structural distortions and degraded prompt alignment. TeaCache alleviates denoising cost through step-wise reuse, but its aggressive reuse can cause unavoidable quality drops. Experimental results demonstrate that under the similar  acceleration ratio, SortBlock consistently outperforms these methods across various evaluation metrics. 

We further evaluate how the quality of generated images degrades under increasing acceleration ratios across different methods. As shown in Fig.~\ref{fig:5}, SortBlock consistently preserves high image quality even at higher acceleration levels.
Notably, images produced by SortBlock exhibit stronger consistency with the baseline model. This is attributed to our targeted recomputation strategy, which selectively refines critical features and mitigates quality loss more effectively than uniform reuse approaches.

\begin{table}[ht]
\centering
\scriptsize
\setlength{\tabcolsep}{2pt}  
\renewcommand{\arraystretch}{1}

\begin{tabular}{lccccccc}
\hline
\textbf{Method} & 
\multicolumn{2}{c}{\textbf{Speed}} & 
\multicolumn{4}{c}{\textbf{Video Quality}} \\
\cmidrule(lr){2-3} \cmidrule(lr){4-7}
& Speedup$\uparrow$ & Lat.(s)$\downarrow$ 
& VBench$\uparrow$ & SSIM$\uparrow$ & PSNR$\uparrow$ & LPIP$\downarrow$ \\
\midrule
Wan2.1 (50 steps) & $1.0\times$ & 86 & 79.55\% & - & - & - \\
\midrule
PAB(5,7,9)             & $1.95\times$ & 44 & 78.83\% & 0.878 & 27.50 & 0.076 \\
TeaCache(rel=0.08)     & $1.95\times$ & 44 & 79.38\% & 0.903 & 28.09 & 0.062 \\
Taylorseer(N=2,O=1)    & $1.79\times$ & 48 & 78.94\% & 0.715 & 18.75 & 0.258 \\
\midrule
SortBlock (K=5)        & \textbf{$2.00\times$} & \textbf{38} 
& \textbf{79.47\%} & \textbf{0.912} & \textbf{29.79} & \textbf{0.048} \\
\hline
\end{tabular}
\caption{Text-to-video generation on Wan2.1.}
\label{tab:wan}
\end{table}

\textbf{Accelerate Wan2.1 and HunyuanVideo for Text-to-Video Generation.} Table~\ref{tab:wan} and Table~\ref{tab:hun} presents a quantitative evaluation of both efficiency and visual quality on Wan2.1 and HunyuanVideo using the VBench benchmark~\cite{huangVBenchComprehensiveBenchmark2023}. Compared with other training-free acceleration baselines, SortBlock consistently achieves superior performance across various backbone models, sampling schedulers, video resolutions, and sequence lengths.

When evaluating on the Wan2.1~\cite{wanteamWanOpenAdvanced2025a} and HunyuanVideo backbone, SortBlock outperforms PAB~\cite{liuReusingForecastingAccelerating2025} and TaylorSeer~\cite{liuReusingForecastingAccelerating2025} in all visual quality metrics under comparable acceleration ratios, and even delivers faster inference than PAB while preserving better image fidelity. Although TeaCache~\cite{liuTimestepEmbeddingTells2025} demonstrates competitive results, SortBlock still maintains a noticeable advantage in terms of both acceleration and visual quality. These results highlight the generalizability and robustness of our method in diverse settings.

\begin{table}[ht]
\centering
\scriptsize
\setlength{\tabcolsep}{2pt}  
\renewcommand{\arraystretch}{1}

\begin{tabular}{lccccccc}
\hline
\textbf{Method} & 
\multicolumn{2}{c}{\textbf{Speed}} & 
\multicolumn{4}{c}{\textbf{Video Quality}} \\
\cmidrule(lr){2-3} \cmidrule(lr){4-7}
& Speedup$\uparrow$ & Lat.(s)$\downarrow$ 
& VBench$\uparrow$ & SSIM$\uparrow$ & PSNR$\uparrow$ & LPIP$\downarrow$ \\
\midrule
HunyuanVideo (50 steps) & $1.0\times$ & 53  & 79.04\% & - & - & - \\
\hline
PAB(4,5,6)            & $1.96\times$ & 27 & 78.06\% & 0.71 & 19.82 & 0.277 \\
TeaCache(rel=0.15)    & $2.40\times$ & 22 & 78.81\% & 0.72 & 20.43 & 0.257 \\
Taylorseer(N=3,O=3)   & $2.40\times$ & 22 & \textbf{79.14\%} & 0.73 & 20.98 & 0.231 \\
\hline
SortBlock (K=5)       & \textbf{$2.40\times$} & \textbf{22} 
& 78.29\% & \textbf{0.85} & \textbf{28.29} & \textbf{0.089} \\
\hline
\end{tabular}
\caption{Text-to-video generation on HunyuanVideo.}
\label{tab:hun}
\end{table}

\textbf{Qualitative Results.} We further conduct qualitative comparisons to illustrate the visual effects of different acceleration strategies, as shown in Fig.~\ref{fig:5}. T-GATE introduces a binary gating mechanism to partition the denoising process into early and late stages. However, this rigid division tends to oversimplify the denoising dynamics, often resulting in blocky artifacts and degraded image quality. Although PAB achieves reasonable visual performance under a 1.37$\times$ speed-up, it requires manual adaptation for each specific Transformer architecture, making it less practical and time-consuming to deploy.

TeaCache achieves favorable acceleration-quality trade-offs by reusing steps in the denoising process. Nevertheless, its coarse-grained reuse strategy can introduce significant structural deviations and semantic distortions, especially in complex scenes. TaylorSeer leverages Taylor series expansion on MLP and attention modules to predict outputs from previous layers. While it effectively captures local historical patterns, it often lacks broader structural awareness, resulting in suboptimal performance on visual metrics.

In contrast, SortBlock offers a 2$\times$ acceleration without requiring any additional training. It selectively recomputes block features based on their predicted similarity, preserving both global structure and fine details. This results in higher fidelity to the original image and eliminates common artifacts observed in other methods.


\subsection{Discussion}
\subsubsection{Ablation Study}
\begin{figure}[ht]
    \centering
    \includegraphics[width=0.95\linewidth]{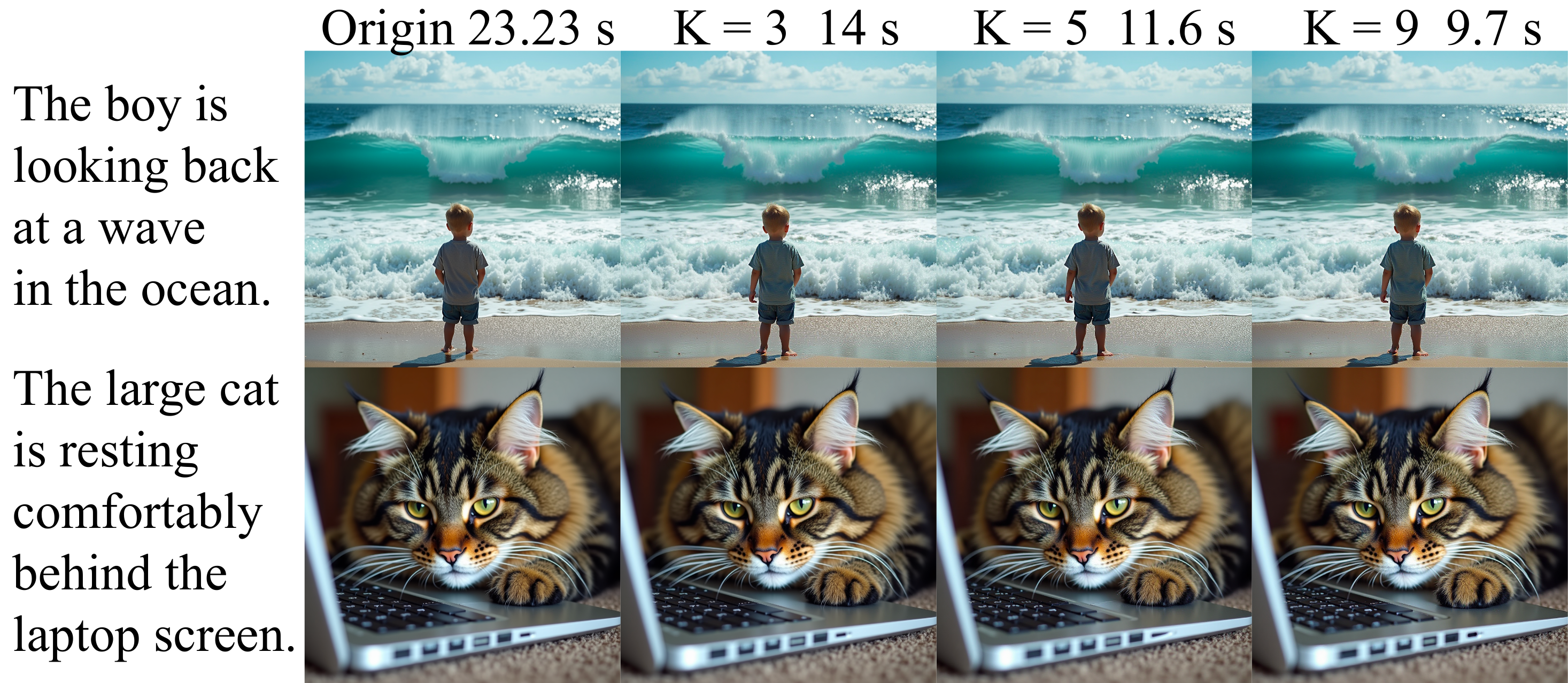}
    \caption{Ablation study on the effect of K.}
    \label{fig:8}
\end{figure}

\textbf{Impact of Reuse Frequency.} As illustrated in Fig.~\ref{fig:8}, we visualize how the generated images evolve with increasing values of $K$. As the generation time decreases, the main semantic content of the image remains consistent, while the fidelity of fine-grained details gradually degrades. This observation is consistent with the insights presented in Table.~\ref{tab:flux}, which models the dynamic recomputation ratio over timesteps. Varying the value of $K$ provides a flexible control mechanism for balancing between generation speed and visual quality.

\begin{figure}[ht]
    \centering
    \includegraphics[width=0.95\linewidth]{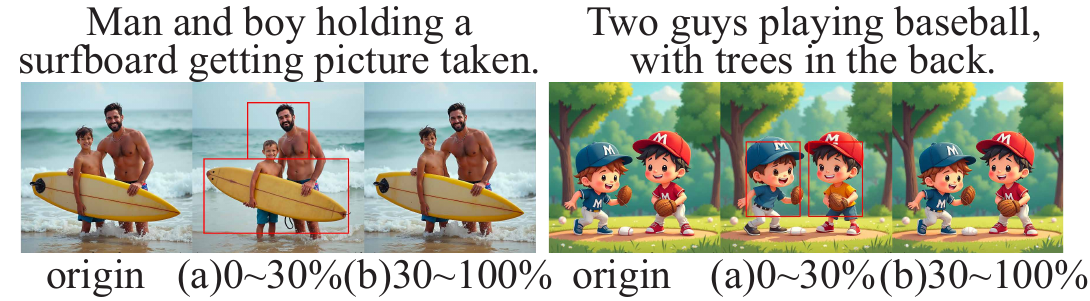}
    \caption{Effect of using SortBlock at different denoising stages.}
    \label{fig:9}
\end{figure}

\begin{figure}[ht]
    \centering
    \includegraphics[width=0.95\linewidth]{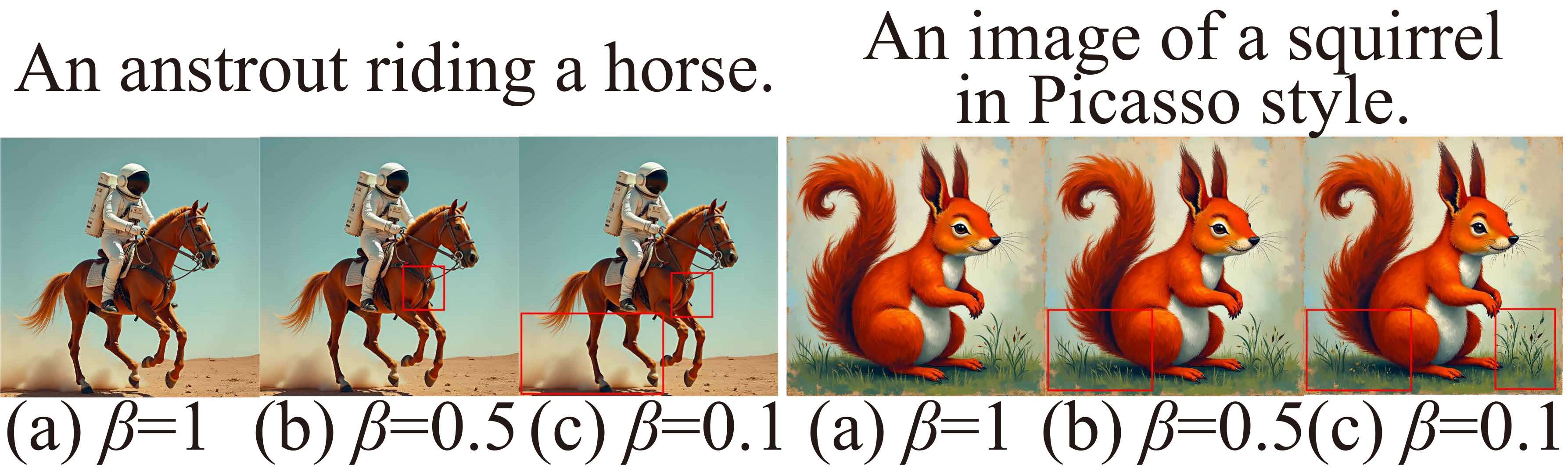}
    \caption{Impact of different $\beta$.}
    \label{fig:10}
\end{figure}

\textbf{Effect of Using SortBlock at Different Denoising Stages.} As shown in Fig.~\ref{fig:9}, we investigate the impact of applying SortBlock at different stages of the denoising process: the early stage (0\%–30\%) and the later stage (30\%–100\%). Since SortBlock primarily reuses blocks with high similarity, applying it during the early stage—where the structural representation is still forming—may introduce structural artifacts or distortions, as highlighted by the red box in Fig.~\ref{fig:9} (a). In contrast, applying SortBlock in the later stage, where the global structure is more stable and the focus shifts to refining texture details, leads to more reliable feature reuse and minimal quality degradation, as illustrated in Fig.~\ref{fig:9} (b). This demonstrates the effectiveness of our adaptive recomputation strategy across different denoising phases.

\textbf{Impact of Reusing Blocks at Different Ratio.} We further investigate the influence of the re-computation ratio by varying the hyperparameter $\beta$, as illustrated in Fig.~\ref{fig:10}. A smaller $\beta$ results in fewer blocks being re-computed, meaning that features in less similar blocks are predicted without explicit recomputation. This leads to a loss of fine-grained details and texture degradation, as highlighted by the red box in Fig.~\ref{fig:10} (b). Conversely, increasing $\beta$ leads to a higher recomputation ratio, preserving more detailed structures in the generated image but at the cost of increased inference time. Therefore, $\beta$ serves as a trade-off knob between image quality and generation efficiency.

\section{Conclusion}
In this paper, we introduced SortBlock , an efficient inference acceleration strategy for diffusion models, which adaptively recomputes block features based on their similarity to full-resolution predictions. By skipping redundant computation for similar blocks, SortBlock significantly reduces inference time without introducing additional training. It is applicable to both image and video generation tasks, achieving over $2\times$ speed-up while maintaining structural and perceptual fidelity. Extensive experiments demonstrate that SortBlock consistently outperforms other training-free acceleration methods across various backbone models, samplers, and resolutions, offering a strong trade-off between speed and quality.
\section{Acknowledgements} This work is supported by the National Key Research and Development Program of China under Grant No. 2023YFB4604100.


\end{document}